# Kinematic Knowledge Maps for Pattern Alignment: Structured Latent Representational Learning in Multimodal Gait Analysis

Dong. Chen, Zonglin. He, and Kenneth MC. Cheung

*Abstract*—**Multimodal clinical AI is limited by weakly aligned inputs and the absence of domain-specific interpretable representations, particularly when learning from dense video stream, structured time-series, and template-based kinematic text. Here we present ScoliDetect, an explainable framework for adolescent idiopathic scoliosis screening from monocular gait video, built around a kinematic knowledge map (KKM) and complementary template-based kinematic text derived from per-sequence pose statics. KKM is a fixed-index structured representation that encodes gait features across absolute motion, self-skeleton configuration and joint-joint signal correlation, providing anchor-referenced multimodal fusion and factor-level interpretation. We integrate video, KKM, and template-based kinematic text through bidirectional cross-attention with latent-bottleneck aggregation. In a multicenter cohort (n = 1,858 after exclusions), prespecified supervised ablations on an external screening cohort show that KKM-mediated multimodal fusion outperforms unimodal models and late concatenation. Under a staged training protocol, trimodal contrastive pretraining is applied after architecture selection as representation initialization, improving external ROC-AUC from 0.961 to 0.972. Furthermore, the structured nature of the KKM provides inherent, factor-level attributions mapped directly to specific kinematic phases and skeletal indices, offering verifiable interpretability. The results demonstrate that embedding explicit structural topologies into latent spaces significantly enhances both the generalization and explainability of multimodal pattern analysis systems.**

We acknowledge funding support by Sanming Project of Medicine in Shenzhen, China (No. SZSM202211004); Shenzhen Science and Technology Program (No. KJZD20240903102759061); Shenzhen-Hong Kong Cooperation Zone for Technology and Innovation (No. HZQSWS-KCCYB-2024055). Corresponding author: Kenneth MC, Cheung.

Dong. Chen is with the Department of Orthopaedics & Traumatology, University of Hong Kong, Hong Kong, China. And the Univerisity of Hong Kong - Shenzhen hospital, Shenzhen, China (e-mail: olichen@connect.hku.hk).

Zonglin. He is with the Department of Orthopaedics & Traumatology, University of Hong Kong, Hong Kong, China. (e-mail: leonhe@connect.hku.hk).

Kenneth MC. Cheung is with the Department of Orthopaedics & Traumatology, University of Hong Kong, Hong Kong, China. And the Univerisity of Hong Kong - Shenzhen hospital, Shenzhen, China (e-mail: cheungmc@hku.hk).

The code is available at: https://github.com/Oli21-chen/ScoliDetect_HKU_SpineSeek_AISscreening

## I. INTRODUCTION

MULTIMODAL machine learning in clinical settings must reconcile heterogeneous sensing streams—raw video, derived time series, and text that are rarely synchronized in time or semantics. When modalities are misaligned, fusion models encode spurious context; when models lack structured intermediates, explanations collapse to opaque pixel saliency. These challenges are acute in periodic biological motion, where gait phase, acquisition protocol, and operator variability jointly determine whether a system generalizes across sites or memorizes deployment context.

Adolescent Idiopathic Scoliosis (AIS) is a three-dimensional spinal deformity affecting 1-5% of adolescents, characterized by a lateral curvature exceeding 10° with vertebral rotation[1-4]. Late detection increases the risk of progression beyond 50°, with substantially higher morbidity and treatment cost than early monitoring[2, 5-7]. Population school screening, long established in Hong Kong and similar programs, relies on forward bending test (FBT), scoliometer measurement of angle of trunk rotation (ATR) and Moiré topography, followed by radiographic confirmation when indicated[8-10] (**Fig. 1a**, right background). Meta-analyses and large cohort studies report meaningful sensitivity but limited positive predictive value and continued reliance on surface deformity, privacy-preserving facilities, trained personnel, and avoidable radiation when referrals escalate[10, 11]. Reported global operating points for conventional school screening modalities are summarized in **Fig. 1b**[10]. Gait kinematics and spinal alignment differ systematically between AIS and controls[12-14], motivating screening approaches that do not require disrobing and can be deployed with brief operator training.

Existing AI approaches of AIS have improved static screening from postural photographs[15, 16], ultrasound[17, 18], smartphone back images[19-21], or Cobb-angle measurement on radiographs[22-24]. These pipelines largely ignore the dynamic gait asymmetries that emerge early in biomechanical compensation[12, 25]. Recent gait-based models use end-to-end video or skeleton graphs[26-28], but they instantiate two recurring limitations in multimodal scientific AI. (i) modality dropping or naive fusion—unimodal pipelines disregard complementary pixels or kinematics, whereas late concatenation fails when streams are temporally misaligned; and (ii) post-hoc explainability—saliency on pixels or joints does not yield stable, named factors tied to model inputs. We hypothesize that when labels are scarce and explanations must be auditable, a predefined representation, such as KKM, can improve both multimodal integration and interpretability relative to end-to-end fusion alone.

In that case, we propose ScoliDetect, an explainable multimodal framework for AIS screening from monocular gait video (**Fig. 1a**) based on kinematic knowledge map (KKM) and its template-based kinematic text (**Fig. 1c**). Radiographic severity in the development cohorts is shown in **Fig. 1d**. This study is built with following three aspects:

**(1) Fixed-index structured latent (KKM).** KKM serializes each aligned gait cycle as a predefined factor grid spanning absolute motion, self-skeleton configuration, and joint–joint signal correlation. The same index set supports frame filtering, template-based kinematic text generation, and factor-level readouts that share coordinates with fusion.

**(2) Supervised fusion with latent-bottleneck aggregation.** Video and KKM are integrated through bidirectional cross-attention and Perceiver-style latent-bottleneck compression. Prespecified supervised ablations on a held-out external screening cohort evaluate unimodal baselines, fusion variants, and late concatenation; trimodal contrastive pretraining is applied only after this architecture is fixed, as encoder initialization rather than as an open-ended contrastive search.

**(3) Temporal registration and structural interpretability.** Gait cycles are aligned by removing stationary frames, synchronizing motion onset, and peak-anchoring on the aggregated KKM trajectory, so video and KKM tokens refer to comparable phases before fusion or pretraining. Because each KKM dimension is domain-tagged, explanations are read out at factor indices and gait phases via encoder temporal attention and attention-weighted gradients; we treat these readouts as attributional correlates, not causal claims.

We evaluate ScoliDetect in a multicenter observational cohort (n = 1,858 after exclusions) with patient-level partitioning, external school screening validation, prespecified ablations, subject-level bootstrap inference.

## II. Related Work

### A. Multimodal Representation and Cross-Modal Fusion

Multimodal learning aligns heterogeneous sensing streams, such as vision, audio, and language, in a shared representation space. Contrastive pretraining aligns paired modalities with InfoNCE-style objectives[29, 30], and recent variants refine image–text alignment at scale[31]. Cross-modal Transformers fuse unaligned token sequences through attention[32], while Perceiver-style architectures compress long inputs through a fixed latent bottleneck before downstream prediction[33]. These designs are effective when modalities are approximately co-registered or when alignment can be learned implicitly from large paired corpora.

Periodic motion in the wild violates these assumptions. Monocular gait video, kinematic knowledge map, and template-based kinematic text are produced by different pipelines, run on different temporal grids, and are sensitive to capture protocol and operator behavior. Under such temporal and semantic misalignment, implicit fusion and contrastive objectives can absorb site-specific acquisition context rather than transferable motion structure. Late concatenation of independently encoded streams[32] further fails when token axes are not comparable across modalities. ScoliDetect treats explicit registration and a shared structured intermediate as first-class design choices, and uses prespecified supervised ablations on a held-out cohort to test fusion hypotheses under fixed training protocols. Contrastive pretraining appears only after the fusion architecture is selected, as encoder initialization rather than as the primary evidence base.

### B. Video, Pose, and Gait Pattern Analysis

Spatiotemporal Transformers are standard encoders for video understanding[34], with rotary or factorized positional schemes supporting long clips[35]. Human motion analysis often derives skeleton sequences from pose estimators[36, 37]. For scoliosis-related gait, primary studies report asymmetry in trunk and limb kinematics[12-14], and meta-analyses summarize altered gait kinematics and lower-limb coordination[12, 25]. Recent computational approaches classify scoliosis from gait video or skeleton sequences[26-28], including dual-representation design and graph convolutional models on skeleton trajectories. These pipelines advance application performance but are predominantly unimodal or fuse streams with limited alignment machinery, and they explain predictions through saliency rather than named factors indexed to predefined model coordinates[16]. On the other hand, Pose-estimation quality itself affects scoliosis screening relative to radiographic reference, reinforcing the need for structured intermediates when pixel-level evidence is noisy.

ScoliDetect constructs a fixed-index kinematic knowledge map from hybrid pose-estimator outputs, uses the KKM trajectory to filter and peak-anchor video frames, and fuses the aligned video and KKM streams through index-biased bidirectional cross-attention with a Perceiver-style latent bottleneck (**Fig. 1c**). Template-based kinematic text, generated from the same per-sequence pose statistics, is encoded independently and concatenated with the pooled video–KKM representation at the classification head, forming the full trimodal model (KVT) without routing text through the video–KKM cross-attention stack.

### C. AI for Scoliosis and Population Screening

Adolescent idiopathic scoliosis affects a substantial fraction of adolescents[1-4]. Population programs rely on forward-bending tests, scoliometer readings, and Moiré topography, with radiographic confirmation when indicated[8-10]. Meta-analyses document meaningful sensitivity but limited positive predictive value at typical prevalence[10, 11]. Deep learning has improved static screening from photographs[15, 16], ultrasound[17, 18], smartphone surface images[19-21], and radiographs[22-24], but these methods use single-time-point anatomy rather than periodic gait compensation. ScoliDetect evaluates monocular walking video on an external school-screening cohort; AIS serves as a stress test for misaligned multimodal video rather than as the primary pattern-analysis contribution. Operator-capture reporting follows early-stage clinical-AI guidance[38].

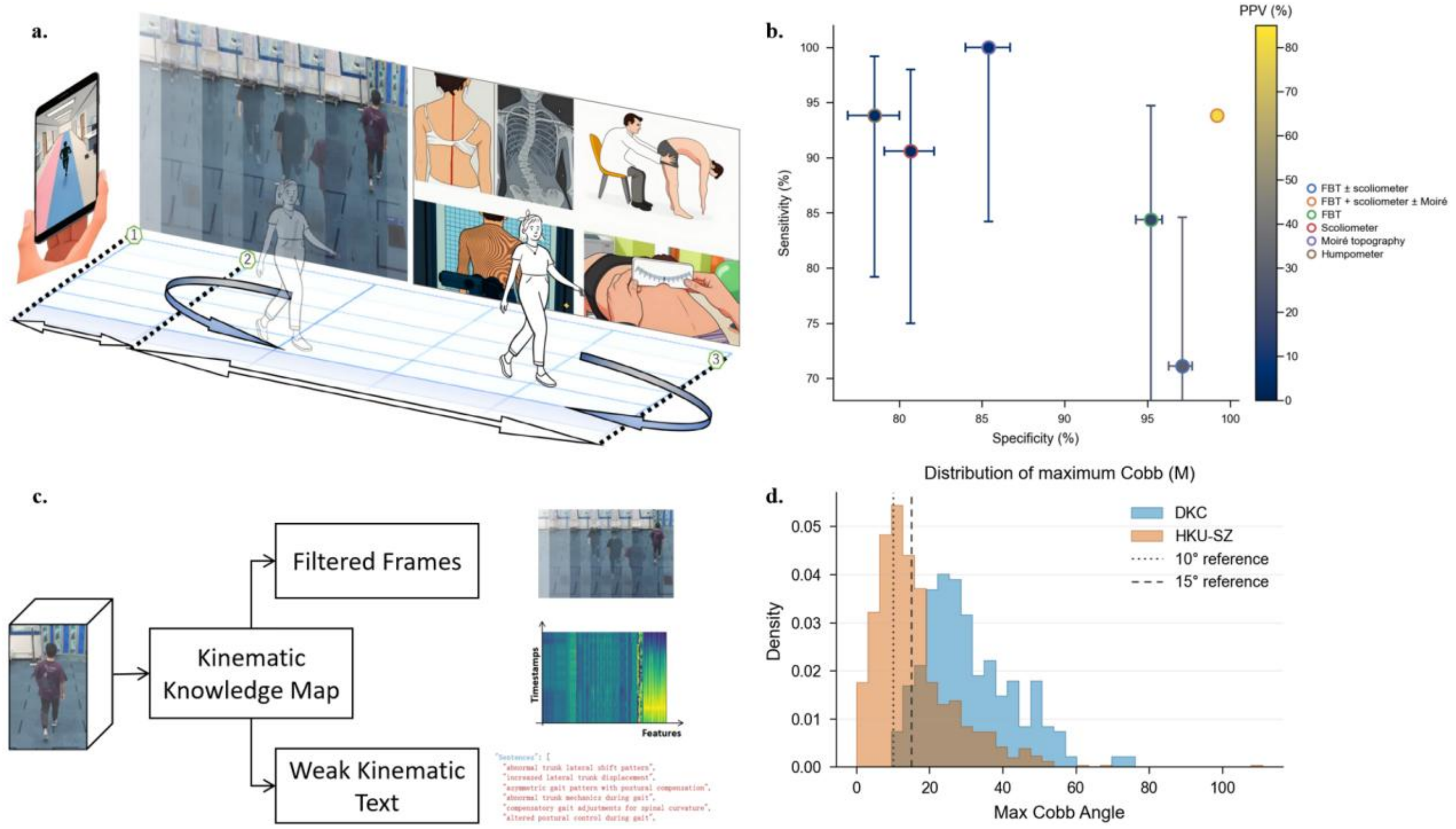

**Fig. 1.** Adolescent idiopathic scoliosis (AIS) screening context. (a) Monocular gait video capture under a standardized walking protocol (left) compared with conventional screening modalities that often require disrobing, specialist examination, or radiation (right): visual inspection, radiography, forward-bending test (FBT), Moiré topography, and scoliometer measurement. (b) Reported sensitivity and specificity of population screening modalities from meta-analyses and cohort studies[10] (points coloured by positive predictive value, PPV). (c) Overview of the kinematic pipeline: gait video is converted to a kinematic knowledge map (KKM) and template-based kinematic text for multimodal learning. (d) Distribution of maximum Cobb angle among participants with radiography (DKC cohort, n = 299; HKU-SZ cohort, n = 820), with reference lines at 10° and 15°.

## III. PROPOSED METHOD

### *A. Overview*

ScoliDetect maps a monocular gait clip to a binary screening logit through two coupled stages: fixed-index representation construction and multimodal fusion with a screening head.

Representation construction (**Fig. 2**). Raw video is converted into aligned inputs for learning: (i) hybrid pose estimation and target-subject selection; (ii) scale-invariant joint normalization and fixed-index KKM construction over motion, skeleton-configuration, and signal-cross-correlation domains; and (iii) peak-anchored temporal registration of video and KKM, together with generation of template-based kinematic text from per-sequence pose statistics.

Multimodal learning (**Fig. 3**). Registered inputs are encoded and fused as follows: (iv) modality-specific encoders for video, KKM, and text; (v) index-biased bidirectional cross-attention with a Perceiver-style latent bottleneck for video–KKM fusion (KV); and (vi) late concatenation of the text embedding with pooled video – KKM features at the screening head (KVT). Factor-indexed interpretability readouts are computed on the KKM grid used by the fusion path. Training objectives and implementation details are given in Sections III-H and IV-D.

### *B. Problem Formulation*

Each sample comprises a monocular gait video clip $\boldsymbol{V}$, an aligned KKM tensor $\boldsymbol{K} \in \mathbb{R}^{T_k \times F}$ with $F$ = 238 predefined factors, and template-based kinematic text $\boldsymbol{T}$ generated from per-sequence pose statistics. After peak-anchored registration, video and KKM share a normalized gait-phase axis but with different resolution.

Factor indices of KKM are fixed before training and partition into absolute motion ($0-63$), self-skeleton configuration ($64-171$), and joint-joint signal correlation ($172-237$) domains; full factor definitions are listed in the supplementary material.

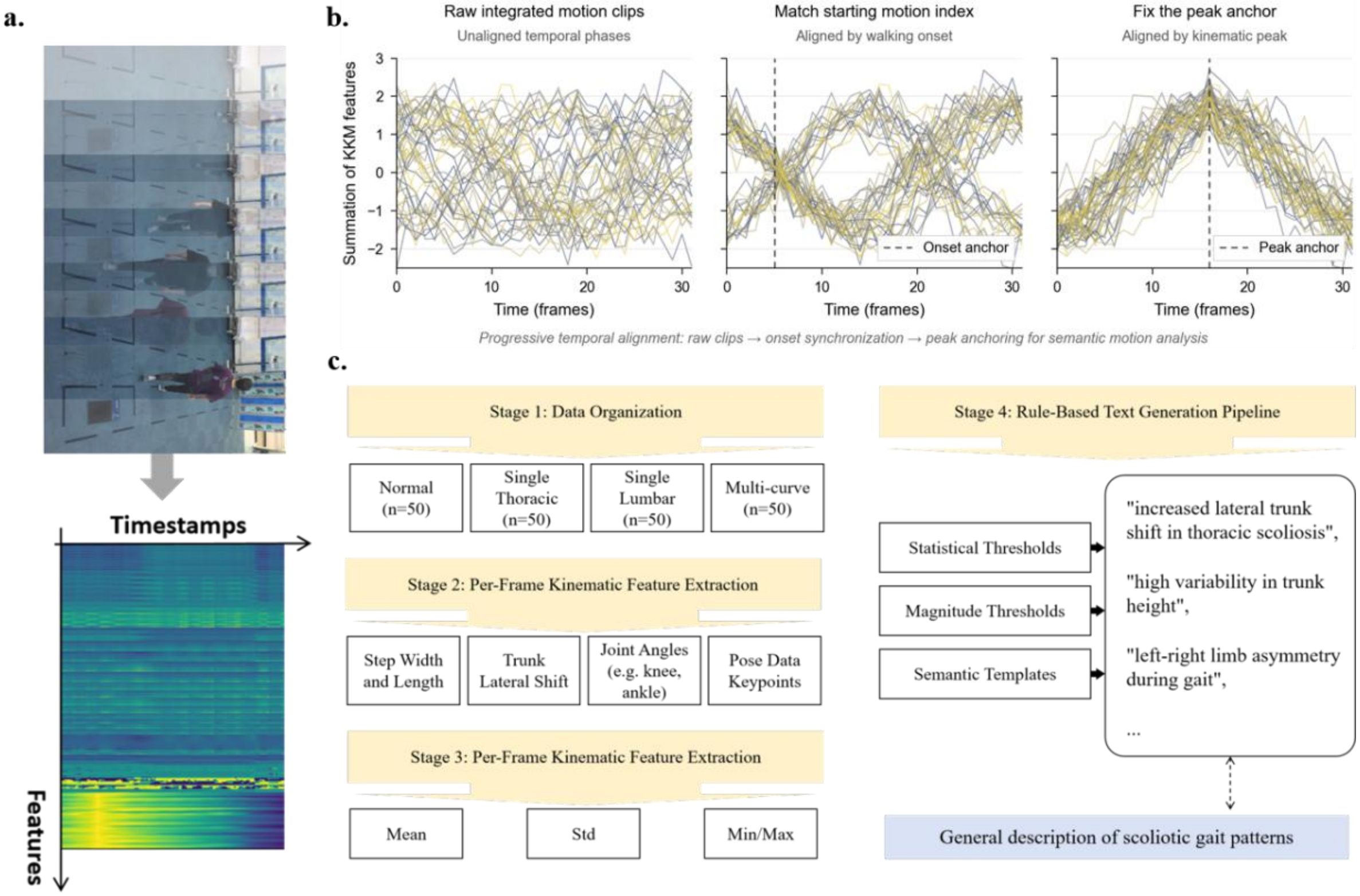

**Fig. 2.** Multimodal data construction and kinematic semantic alignment. (a) Hybrid pose estimation and fixed-index KKM construction. (b) Progressive temporal registration from onset to peak anchor. (c) template-based kinematic text generation from per-sequence statistics.

*C. Pose Estimation and Target Selection*

Joint landmarks are extracted with a hybrid bottom-up and top-down pose estimator[36, 37]. The bottom-up branch detects all joints and groups them into skeletons, improving robustness to occlusion; the top-down branch detects each person and estimates joints per identity, improving temporal consistency. In multi-person scenes, the target subject is the candidate closest to the central screen, matched from bottom-up proposals to a top-down reference skeleton (**Fig. 2a**).

In multi-person scenes, we select the target subject closest to the central screen by matching bottom-up candidates to the top-down reference. Let the top-down estimator produce a reference joint set $\boldsymbol{j}_k$, where $\boldsymbol{j}_k$ denotes the pixel coordinates of the $k^{th}$ joint. Let the bottom-up estimator yields $\boldsymbol{s}_{i,k}$ candidate skeletons, the target skeleton is selected as

$$\mathcal{S}^* = \underset{\mathcal{S}_i}{argmin}\frac{1}{K}\sum_{k=1}^{K} \| \boldsymbol{s}_{i,k} - \boldsymbol{j}_k \|_2^2 \tag{1}$$

*D. Scale-Invariant Normalization and KKM Construction*

Raw 2D joint coordinates from monocular frontal video depend on camera height pose and subject distance. Each frame is re-centered on the hip midpoint, scaled by a torso-derived factor, and divided by the maximum joint distance to obtain scale-invariant trajectories (**Fig. 2b**).

For each frame $t$, let $\mathbf{P}_t = \{\mathbf{p}_{t,k}\}_{k=1}^{K}$ be the raw pixel coordinates of skeletal keypoints. The scale normalization proceeds in three steps:

(S1) Origin transformation. The subject-specific origin is the midpoint of the hip:

$$\mathrm{o_t} = \frac{1}{2}\left(\mathrm{p_{t,Lhip}} + \mathrm{p_{t,Rhip}}\right) \tag{2}$$

and all joints are shifted to this origin:

$$\tilde{\mathrm{p}}_{\mathrm{t,k}} = \mathrm{p_{t,k}} - \mathrm{o_t} \tag{3}$$

(S2) Dynamic pose scaling. To compensate for variable subject-camera distance, we compute a per-frame scale factor $\mathrm{s_t}$. First, the torso size is defined as the Euclidean distance between the shoulder centre and the hip centre:

$$\tau_t^{\text{torso}} = \|\mathbf{c}_t^{\text{shoulder}} - \mathbf{c}_t^{\text{hip}}\|_2 \tag{4}$$

Then, the maximum distance from any joint to the origin is:

$$r_t = \max_k \|\tilde{\mathbf{p}}_{t,k}\|_2 \tag{5}$$

The final pose scale is:

$$s_t = \max\left(2.5\tau_t^{\text{torso}},\ r_t\right) \tag{6}$$

where the multiplier 2.5 follows the heuristic from BlazePose[36] to ensure that even when the subject is far from the camera, the scale is not dominated by a single outlier joint.

(S3) Coordinate normalization. Finally, all origin-shifted coordinates are divided by the pose scale:

$$\widehat{\mathbf{p}}_{t,k} = \widehat{\mathbf{p}}_{t,k}/s_t \tag{7}$$

The resulting $\widehat{\mathbf{p}}_{t,k}$ are invariant to camera height pose and subject distance while preserving pathological asymmetries (e.g., pelvic tilt or shoulder height difference) because the same scale factor is applied to all joints within a frame. These normalized trajectories are used as the input for all subsequent KKM feature computations.

The KKM operationalizes a structured-latent hypothesis: holistic gait can be approximated by a fixed dictionary of interpretable kinematic factors. Full factor definitions and formulas are provided in the **Supplementary method section**. Domain-wise magnitude normalization is applied before model input formatting.

### *E. Temporal Alignment and Kinematic Text*

Video and KKM streams are temporally registered using the summed KKM trajectory (**Fig. 2c**). Stationary leading frames are removed; remaining sequences are coarse-aligned by motion onset and fine-aligned by peak anchoring of the summed KKM curve. template-based kinematic text is generated from per-sequence pose statistics (mean and standard deviation of step width, trunk shift, and joint angles) using category-specific templates (normal, single thoracic, single lumbar, multi-curve); template rules are given in the supplementary material.

### *F. Modality-Specific Encoders*

Each modality is mapped to a shared hidden dimension before fusion (**Fig. 3a**). The video encoder uses a Video Vision Transformer (ViViT)[34] on 32 × 224 × 224 clips with 1 × 16 × 16 spatiotemporal voxels and rotary position embeddings (RoPE)[35]. The KKM encoder is a standard Transformer[34] with RoPE on temporal tokens. Text is encoded with a pretrained Sentence Transformer (all-MiniLM-L6-v2)[39] followed by a two-layer MLP projection (LayerNorm in pretraining, dynamic Tanh normalization during fine-tuning [40].

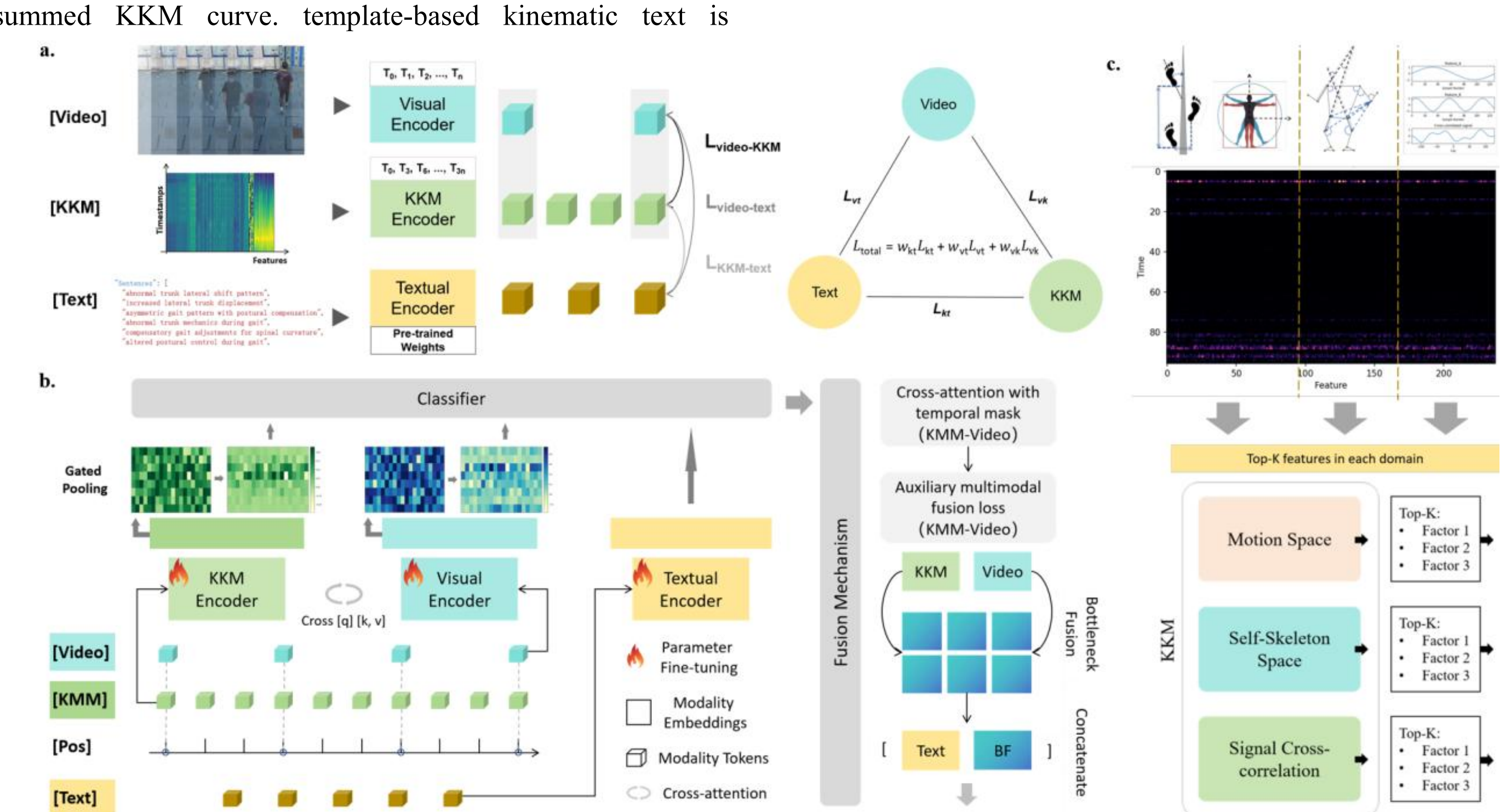

**Fig. 3.** Multimodal learning: contrastive alignment, supervised fusion, and structural explanation. (a) Trimodal contrastive pretraining (InfoNCE). (b) Supervised KVT head (cross-attention, Perceiver-style fusion, late text concat). (c) Fixed-index KKM domains with top-k readouts.

### *G. Multimodal Fusion Architecture*

Fusion couples video and KKM through index-biased bidirectional cross-attention and a Perceiver-style latent bottleneck[33, 41]. Template-based kinematic text does not enter the video-KKM cross-attention stack; $\mathbf{z}_\tau$ is late-concatenated at the classification head (KVT), as shown in **Fig. 3b**.

For each mini-batch sample, one-dimensional index tracks are provided for video and KKM token sequences $\boldsymbol{i}^v \in \mathbb{R}^{T_v}$ and $\boldsymbol{i}^k \in \mathbb{R}^{T_k}$ ,where $T_v$ and $T_k$ are the numbers of tokens entering cross-attention. This allows coarse frame-level or clinical indices to be registered with encoder token grids.

To remove absolute scale differences between modalities while preserving within-sequence ordering, each resampled track is min–max normalized to [0, 1], yielding $\bar{\boldsymbol{i}}_t^v$ and $\bar{\boldsymbol{i}}_s^k$. Pairwise absolute differences define a distance matrix $\boldsymbol{D}^{v\to k} \in \mathbb{R}^{T_v \times T_k}$. The distances are converted into an additive attention bias that softly encourages attention between temporally co-located tokens:

$$\boldsymbol{B}^{v\to k} = -\lambda \boldsymbol{D}_{t,s}^{v\to k} = -\lambda \left| \bar{\boldsymbol{i}}_t^v - \bar{\boldsymbol{i}}_s^k \right| \quad (11)$$

$$\boldsymbol{B}^{k\to v} = \left(\boldsymbol{B}^{v\to k}\right)^\top \quad (12)$$

where $\lambda \geq 0$ controls bias strength. Bias tensors are broadcast over heads to match multi-head attention layout and added to raw attention logits before softmax in the corresponding cross-attention blocks.

Inspired from the previous studies[33, 41], we maintain a learnable latent array $\boldsymbol{Z_0} \in \mathbb{R}^{M\times d}$, with $M$ bottleneck tokens and hidden size $d$. After the bidirectional video-KKM cross-attention stack, we denote the resulting token sequences by $\boldsymbol{X_v} \in \mathbb{R}^{T_v\times d}$ and $\boldsymbol{X_k} \in \mathbb{R}^{T_k\times d}$. For each bottleneck layer $\ell$, we update the latents in two cross-attention steps: latent queries attend to video tokens as keys/values, then the same latent queries attend to KKM tokens as keys/values,

$$\widetilde{\boldsymbol{Z}}^{\ell} = CA_v^{\ell}\left(\boldsymbol{Z}^{\ell-1}, \boldsymbol{X_v}; \boldsymbol{B}^{k\to v}\right), \quad \boldsymbol{Z}^{\ell} = CA_k^{\ell}\left(\widetilde{\boldsymbol{Z}}^{\ell}, \boldsymbol{X_k}; \boldsymbol{B}^{v\to k}\right) \quad (13)$$

where $CA_v^{\ell}$ and $CA_k^{\ell}$ are multi-head cross-attention blocks with latent queries and modality tokens as keys and values. Unlike the main video–KKM cross-attention path, these bottleneck blocks do not apply the temporal resolution index bias. The fused bottleneck summary used downstream is obtained by mean-pooling over the latent tokens after the last layer, followed by a modality-specific normalization block:

$$\bar{\boldsymbol{z}} = Norm\left(\frac{1}{M}\sum_{m=1}^{M} \boldsymbol{Z}_{\boldsymbol{m}}^{\ell}\right) \quad (14)$$

In parallel for the two modalities, each sequence $\boldsymbol{X_v}$ and $\boldsymbol{X_k}$ is reduced to a single vector by gated token pooling, then passed through modality-specific normalization:

$$\widetilde{\boldsymbol{x}}_{\boldsymbol{v}} = Norm_v\left(Pool_v(\boldsymbol{X_v})\right), \quad \widetilde{\boldsymbol{x}}_{\boldsymbol{k}} = Norm_k\left(Pool_k(\boldsymbol{X_k})\right) \quad (15)$$

The shared bottleneck vector $\bar{\boldsymbol{z}}$ is mapped with learned linear maps and added to both modality sequences:

$$\widetilde{\boldsymbol{z}}_{\boldsymbol{v}} \leftarrow \widetilde{\boldsymbol{x}}_{\boldsymbol{v}} + W_v\,\bar{\boldsymbol{z}}, \quad \widetilde{\boldsymbol{z}}_{\boldsymbol{k}} \leftarrow \widetilde{\boldsymbol{x}}_{\boldsymbol{k}} + W_k\bar{\boldsymbol{z}} \quad (16)$$

These vectors $\widetilde{\boldsymbol{z}}_{\boldsymbol{v}}$ and $\widetilde{\boldsymbol{z}}_{\boldsymbol{k}}$, are then used together with the text branch $\widetilde{\boldsymbol{z}}_{\boldsymbol{t}}$ for the final classifier input.

$$\boldsymbol{h} = [\widetilde{\boldsymbol{z}}_{\boldsymbol{v}}; \widetilde{\boldsymbol{z}}_{\boldsymbol{k}};\ \widetilde{\boldsymbol{z}}_{\boldsymbol{t}}] \in \mathbb{R}^{d_v+d_k+d_t} \quad (17)$$

This vector is processed by a two-layer multilayer perceptron (MLP) with intermediate normalization, nonlinearity and dropout regularization:

$$\boldsymbol{u} = Dropout\left(\phi\left(Norm(\boldsymbol{W_1}\boldsymbol{h} + b_1)\right)\right) \quad (18)$$

$$y = \boldsymbol{W_2}\boldsymbol{u} + b_2 \quad (19)$$

In our implementation, Norm is layer normalization and $\phi$ is ReLU activation function. The scalar output $y$ is the classification logit for the binary endpoint.

*H. Optimization Objectives*

Training is staged: trimodal contrastive pretraining aligns video, KKM, and text in a shared 512-dimensional space using symmetric InfoNCE losses[29, 31] on all modality pairs; this stage initializes encoders only after the fusion architecture is fixed by supervised ablations. Supervised fine-tuning uses focal loss[42] for class imbalance and an optional symmetric video–KKM InfoNCE term weighted by $\lambda$ (Equations below). Scratch baselines randomize all weights under the same schedule. Optimizer settings and learning-rate schedules are specified in Section IV-D.

For a batch of N triplets $\{(\mathrm{V_i}, \mathrm{K_i}, \mathrm{T})\}_{\mathrm{i}=1}^{\mathrm{B}}$, let $\mathrm{z}_{\mathrm{i}}^{(\mathrm{a})}$ denote the $\ell_2$-normalized embedding of modality a. Pairwise similarity matrices are

$$S_{ab}[i,j] = \left(z_i^{(a)}\right)^{\top} z_j^{(b)}/\tau_{temp} \quad (20)$$

where $\tau_{\mathrm{temp}}$ is a learnable temperature. The symmetric CLIP-style[29, 31] objective is applied to each modality pair; the total pretraining loss is

$$\mathcal{L}_{pre} = \mathcal{L}_{vt} + \mathcal{L}_{vk} + \mathcal{L}_{kt} \quad (21)$$

For supervised fine-tuning, let $y_i \in \{0,1\}$ be the ground-truth label and $p_i = \sigma(\hat{y}_i)$ be the predicted probability, the focal loss is

$$\mathcal{L}_{focal} = -\frac{1}{N}\sum_{i=1}^{N} w_{t,i}\left(1 - p_{t,i}\right)^{\gamma} log(p_{t,i}), \quad (22)$$

where $w \in [0,1]$ controls class weighting and $\gamma \geq 0$ is the focusing parameter. An optional auxiliary symmetric InfoNCE[29, 30] term $\mathcal{L}_{\mathrm{aux}}$ over projected video and KKM features yields

$$\mathcal{L} = \mathcal{L}_{focal} + \lambda\mathcal{L}_{aux} \quad (23)$$

where $\lambda_{align} \geq 0$ controls the strength of auxiliary multimodal alignment.

*I. Factor-level Interpretability*

Interpretability is structural on the KKM grid and attributional at inference (**Fig. 3c**). Each input dimension maps to a predefined factor index; native KKM temporal attention and attention-times-gradient maps rank factors and gait phases for a fixed classifier. These readouts indicate correlates of the prediction, not causal mechanisms.

The attribution map is

$$\mathcal{A}(t,f) = \alpha_t \cdot \left|\frac{\partial y}{\partial K_{t,f}}\right| \quad (24)$$

where $\mathrm{K_{t,f}}$ is the KKM input at time t and feature f, y is the screening logit, and $\alpha_{\mathrm{t}}$ is the normalized temporal attention from the KKM encoder. Top-*K* factors are reported per domain (motion, self-skeleton, signal cross-correlation).

## IV. EXPERIMENTAL SETUP

*A. Ethics and Participants*

AIS gait screening was conducted at The Duchess of Kent Children's Hospital (DKC) between October 2021 and December 2022; the University of Hong Kong - Shenzhen Hospital (HKU-SZ) between July 2022 and November 2025; and Shenzhen Hong Kong Pui Kiu College Longhua Xinyi School (Pui Kiu) between May 2024 and July 2024. This study was approved by the Institutional Review Board of The University of Hong Kong/Hospital Authority Hong Kong West Cluster (Ref No.: UW 21-511) and the University of Hong Kong - Shenzhen Hospital Review Board (Ref No. Hkuszh2023020). All participants or their guardians provided written informed consent after explanation of study purpose and procedures. Inclusion criteria was independent ambulation without assistive devices. Exclusion criteria were documented gait impairment, other musculoskeletal disorders affecting walking, prior mobility-related surgery, or severe lower- or upper-limb deformities that could confound gait analysis.

Nine non-specialist operators (four medical students, two engineering doctoral candidates, two nurses, one school teacher) completed data collection in a hospital gait laboratory, outpatient corridors, and school classrooms. After a standardized briefing, median hands-on capture time was 30 min per operator (range 10–20 min training; 87 independent sessions: medical students, n = 15; engineering students, n = 44; nurses, n = 20; school teacher, n = 8). Checklist-based protocol adherence was high for tripod placement, recording start, and upload; minor lighting or positioning deviations did not prevent successful capture, and no operator required retraining.

### *B. Data Collection*

Gait videos were acquired at 1080p and 30 fps using a tripod-mounted camera at 1.5 ± 0.1 m height under a standardized walking protocol. Protocol adherence was assessed per session with a fixed checklist: (1) correct tripod height 4 ± 0.1m, (2) subject visible full torso on screen, (3) recording started after ≥ 1s still standing, (4) continuous walk without pause, (5) successful upload.

### *C. Data Partition*

At DKC, 37 participants without radiographic measurements were excluded, leaving 299 (Dataset 1) for external evaluation. At HKU-SZ, 79 without radiographs were excluded, leaving 820 (Dataset 2), split into training (n = 656) and validation (n = 164). Among school participants, 584 formed Dataset 3 (school screening cohort) and 155 formed Dataset 4 (PK): pre-selected controls with radiographic confirmation of non-scoliosis (Cobb ≤10°), used together with DK non-AIS cases as shared negative set (n = 172) in external subgroup and screening analyses. After exclusions, 1,858 participants contributed to model development and evaluation (**Fig. 4; Table I**). Pretraining combined Datasets 2 and 3 (n = 1,404); the external test set combined Datasets 1 and 4 (n = 454).

### *D. Subgroup Method*

ROC-AUC confidence intervals and pairwise subgroup contrasts used participant-level bootstrap resampling (2,000 replicates, fixed seed where applicable). For each replicate, participants were resampled with replacement and AUC was recomputed on the induced sample set. Paired ΔAUC between subgroups used the same bootstrap draw for both strata. Multiplicity adjustment across pairwise comparisons used the Holm procedure.

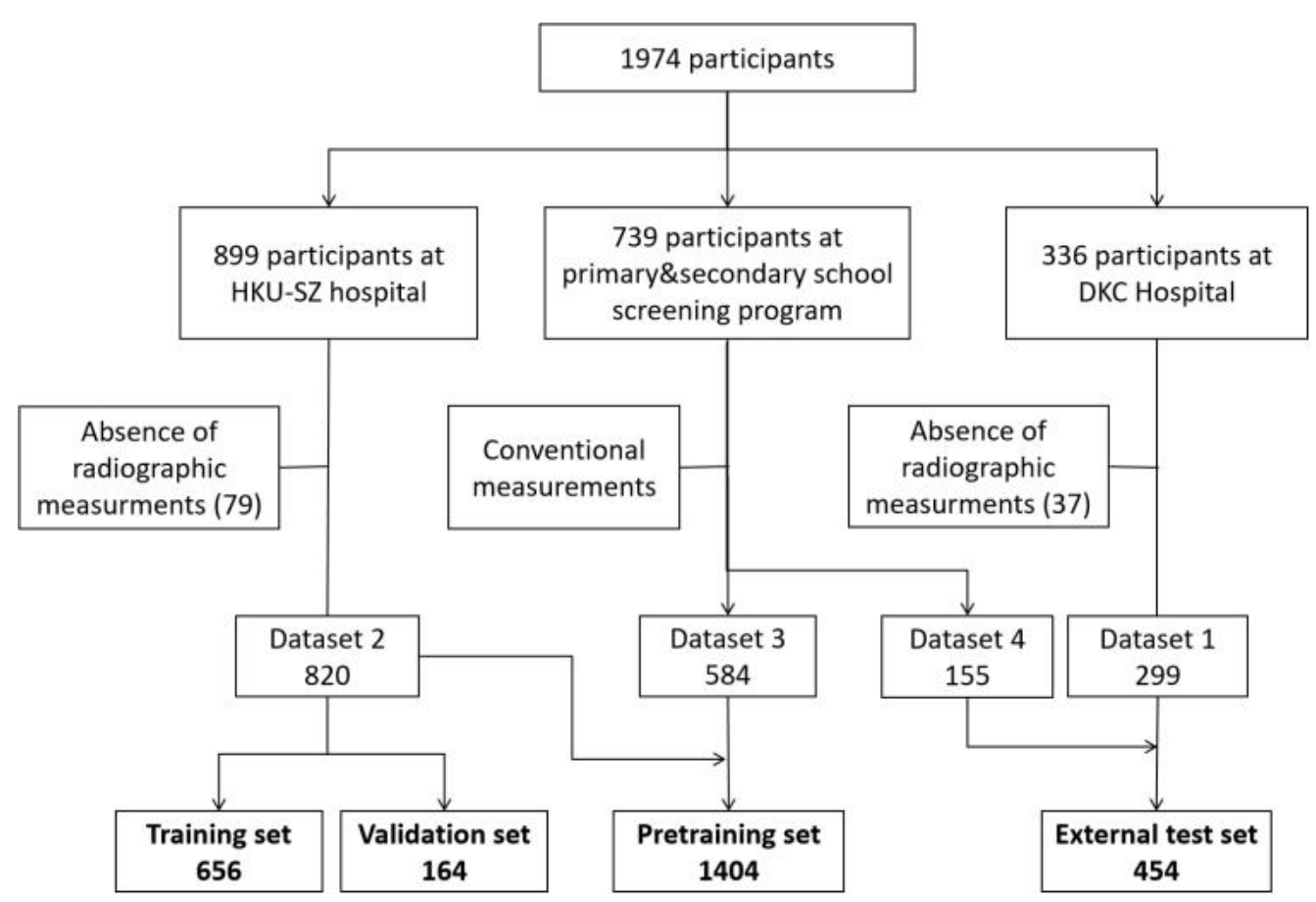

**Fig. 4.** Multicenter recruitment and participant-level data partitioning (enrolled n = 1,974; analyzed n = 1,858). Datasets 1–4 defined by analytic role.

**TABLE I**
DEMOGRAPHIC CHARACTERISTICS OF MULTI-CENTER DATASETS

| Attribute | DKC | HKU-SZ | School |
|---|---|---|---|
| Participants, n | 299 | 820 | 739 |
| Gender (F:M) | 213:86 | 531:289 | 412:327 |
| Age (mean ± SD, y) | 14.97 ± 2.23 | 14.58 ± 6.25 | — |
| Cobb angle (mean ± SD, °) | 30.83 ± 12.31 | 15.63 ± 11.04 | — |
| AIS:Non-AIS | 282:17 | 523:297 | 155 controls |

a PK school controls (n = 155) had radiographic confirmation of Cobb ≤ 10° and contribute all PK negatives in external evaluation, pooled with DKC non-AIS samples as shared negatives (n = 172).

### *E. Training Details*

All models were implemented in PyTorch and trained on NVIDIA A100 (40 GB) GPUs, with random seeds fixed to 42 for reproducibility. The AdamW optimizer was used with β1= 0.9, β2= 0.98, weight decay=0.1. A cosine annealing learning rate schedule with 5% linear warmup was applied in all experiments. For trimodal pretraining, the batch size was 64 and the model was trained for 200 epochs, using early stopping if the validation loss did not improve for 20 consecutive epochs. The learning rate was initialized at $1\times10^{-4}$ and decayed to a minimum of $1\times10^{-6}$. For supervised fine-tuning (SFT) optimization, we used a lower learning rate range ($8\times10^{-6}$ to $1\times10^{-7}$) to stabilize downstream adaptation, with a batch size of 32 and up to 100 epochs (early stopping patience of 15). Scratch baselines followed the same schedule as SFT but with all weights randomly initialized.

In pretraining, the multimodal backbone was optimized to learn transferable cross-modal representations before any task-specific supervision. During SFT, pretrained encoder weights initialized the downstream model, and we employed a staged optimization procedure. First, only the task - specific heads were trained for 20 epochs while all encoders remained frozen. Second, the last two encoder blocks of each modality were unfrozen, and joint training continued for 30 epochs (epochs 21 – 50). Finally, all remaining encoder blocks were unfrozen and the entire model was fine-tuned jointly for the remaining 50 epochs (epochs 51–100).

## V. EXPERIMENTAL RESULTS

### *A. Overall Screening Performance*

On the hold-out external screening cohort (n = 454), the selected KVT model reached ROC-AUC 0.972 after trimodal encoder pretraining and supervised fine-tuning. Prespecified supervised ablations on the same cohort (**Fig. 5a; Table II**)—not backbone choice—provide the primary evidence for multimodal design; encoder comparisons are reported as robustness checks.

Video-only, KKM-only, KV, and KVT variants exhibit a monotonic gain in external AUC (0.784 ⟶ 0.927 ⟶ 0.947 ⟶ 0.961); pretraining improves KVT to 0.972. With modalities and pretraining fixed, latent-bottleneck fusion exceeds late concatenation followed by an MLP (0.972 vs. 0.967, $\Delta$AUC = +0.005). Component ablation on the fixed KVT architecture (**Fig. 5b; Table III**) showed contributions from auxiliary video–KKM InfoNCE, bottleneck fusion, bidirectional cross-attention, and temporal-alignment features; removing temporal alignment reduced performance to 0.943.

**TABLE II**
MODALITY ABLATION ON THE HOLD-OUT TEST SET (N = 454)

| Variant | Pretraining | AUC |
|---|---|---|
| Video only | -- | 0.784 |
| KKM only | -- | 0.927 |
| KV (video + KKM) | -- | 0.947 |
| KVT + concatenate | No | 0.961 |
| KVT + concatenate | Yes | 0.967 |
| KVT + latent-bottleneck | Yes | **0.972** |

**TABLE III**
COMPONET ABLATION OF THE KVT MODEL ARCHITECTURE

| Configuration | AUC |
|---|---|
| KVT+ concatenate (w/o pretrained) | 0.961 |
| w/o auxiliary video–KKM InfoNCE | 0.957 |
| w/o bottleneck fusion | 0.950 |
| w/o bidirectional video–KKM cross-attention | 0.949 |
| w/o phase-index / temporal alignment | 0.943 |

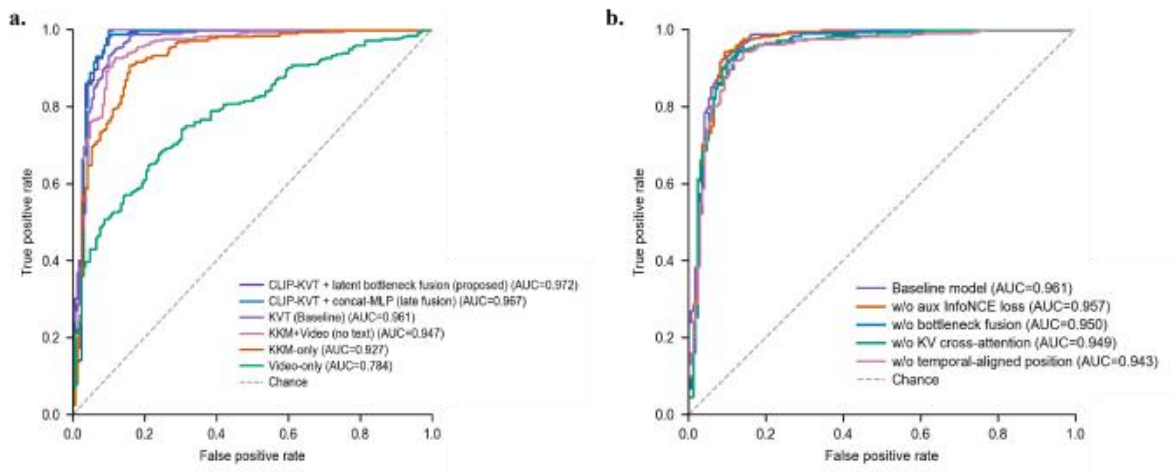

**Fig. 5.** External ROC curves (n = 454). (a) Modality/fusion ablation. (b) Component ablation. AUC values in Tables II and III.

### *B. Backbone Robustness*

On internal validation, Transformer-based backbones showed comparable discrimination and consistently outperformed pretrained ResNet variants. On the external test set, all Transformer models improved relative to their internal validation performance, with ViViT (scratch) reaching an AUC of 0.961. The best overall result was obtained by ViViT with trimodal pretraining followed by supervised fine-tuning, achieving an external AUC of 0.972 (**Table IV**).

The internal-to-external performance shift likely reflects cohort heterogeneity rather than backbone instability. Internal validation contains a higher proportion of self-referred hospital cases, whereas external evaluation is dominated by protocol-driven school screening data (DKC Dataset 1 and PK controls) acquired under more standardized conditions. This acquisition consistency is associated with a more stable gait phenotype distribution and improved external ranking performance.

**TABLE IV**
VIDEO ENCODER BACKBONE COMPARISON

| Visual backbone | Internal val. AUC | External test AUC |
|---|---|---|
| R(2+1)D-18 (pretrained) | 0.751 | 0.377 |
| R3D-18 (pretrained) | 0.743 | 0.641 |
| Video Swin | 0.813 | 0.913 |
| TimeSformer | 0.826 | 0.924 |
| ViViT | 0.787 | 0.961 |
| ViViT (pretrained) | 0.656 | **0.972** |

### *C. Subgroup Evaluation*

We tested whether external performance varied across Cobb-angle severity and curve phenotype on the external test set. Each stratum reused the same shared-negative set, comprising all DKC non-AIS patches union all PK control patches (n = 172 negatives). Uncertainty was estimated with subject-level bootstrap (2,000 resamples). Subgroup contrasts used paired bootstrap ΔAUC with Holm adjustment for multiplicity. Because screening prevalence is low, we also report fixed high-sensitivity operating points (sensitivity ≥ 0.95) with projected PPV, NPV, and false positives per 1,000 screened at 5% assumed prevalence.

a) Severity subgroups: Severity was defined as mild (10° ≤ Cobb < 25° ), moderate (25° ≤ Cobb ≤ 40° ), and severe (Cobb > 40°) among DKC AIS-positive samples in the subgroup file. ROC-AUC was 0.972 [0.952–0.989] (**Table V**). Pairwise ΔAUC values were near zero, all bootstrap intervals included zero, and Holm-adjusted p-values were 1.0 (**Table VI**). At sensitivity ≥ 0.95, specificity ranged from 0.919 to 0.930, with projected false positives per 1,000 screened of roughly 66–77; projected PPV was about 0.38–0.42 and projected NPV ≥ 0.997 (**Table VII**).

**TABLE V**
CURVE SEVERITY SUBGROUP WITH UNCERTAINTY

| Group | N (pos/neg) | AUC | 95% CI |
|---|---|---|---|
| Mild | 274 (102/172) | 0.9725 | [0.9521, 0.9888] |
| Moderate | 288 (116/172) | 0.9711 | [0.9507, 0.9879] |
| Severe | 236 (64/172) | 0.9746 | [0.9551, |

0.9897]

Mild: 10° ≤ Cobb < 25°; moderate: 25° ≤ Cobb ≤ 40°; severe: Cobb > 40° among DKC AIS-positive samples.

TABLE VI
PAIRWISE ΔAUC OF CURVE SEVERITY GROUPS

| Comparison | ΔAUC | 95% CI | p (raw) | p (Holm) |
|---|---|---|---|---|
| Mild − Moderate | 0.0013 | [−0.0056, 0.0088] | 0.726 | 1 |
| Mild − Severe | −0.0021 | [−0.0109, 0.0053] | 0.565 | 1 |
| Moderate − Severe | −0.0034 | [−0.0124, 0.0042] | 0.350 | 1 |

TABLE VII
CURVE SEVERITY STRATA AT SENSITIVITY TARGET ≥ 0.95

| Group | Threshold | Sensitivity | Specificity | PPV (5%) | NPV (5%) |
|---|---|---|---|---|---|
| Mild | 0.4274 | 0.951 | 0.9302 | 0.4177 | 0.9972 |
| Moderate | 0.4184 | 0.9569 | 0.9186 | 0.3822 | 0.9975 |
| Severe | 0.4281 | 0.9688 | 0.9302 | 0.4222 | 0.9982 |

PPV and NPV projected at 5% assumed screening prevalence.

b) Curve-type subgroups: Curve-type labels are defined at the cases level. ROC-AUC was 0.972 [0.953-0.989] (**Table VIII**). Pairwise ΔAUC estimates were small and non-significant after Holm adjustment (**Table IX**). At target sensitivity ≥ 0.95, subgroup-specific specificity ranged from 0.901 to 0.930, with projected false positives per 1,000 screened between approximately 66 and 94; projected PPV ranged from about 0.35 to 0.42, and projected NPV was ≥ 0.997 (**Table X**). Thus the numerically low projected PPV values reflect the chosen prevalence (5%) and sensitivity-first policy, not a collapse of model ranking performance within strata.

TABLE VIII
CURVE TYPE SUBGROUP WITH UNCERTAINTY

| Group | N (pos/neg) | AUC | 95% CI |
|---|---|---|---|
| Overall | 454 (282/172) | 0.9724 | [0.9527, 0.9890] |
| Single thoracic | 222 (50/172) | 0.9701 | [0.9486, 0.9871] |
| Single lumbar | 198 (26/172) | 0.9691 | [0.9455, 0.9876] |
| Multi-curve | 378 (206/172) | 0.9734 | [0.9537, 0.9899] |

TABLE IX
PAIRWISE ΔAUC OF CURVE TYPE GROUPS

| Comparison | ΔAUC | 95% CI | p (raw) | p (Holm) |
|---|---|---|---|---|
| Overall − Single thoracic | 0.0023 | [−0.0044, 0.0108] | 0.508 | 1 |
| Overall − Single lumbar | 0.0033 | [−0.0061, 0.0156] | 0.549 | 1 |
| Overall − Multi-curve | −0.0010 | [−0.0033, 0.0009] | 0.315 | 1 |
| Single thoracic − Single lumbar | 0.0010 | [−0.0125, 0.0150] | 0.892 | 1 |
| Single thoracic − Multi-curve | −0.0032 | [−0.0135, 0.0053] | 0.463 | 1 |
| Single lumbar − Multi-curve | −0.0042 | [−0.0172, 0.0072] | 0.479 | 1 |

TABLE X
CURVE TYPE STRATA AT SENSITIVITY TARGET ≥ 0.95

| Group | Threshold | Sensitivity | Specificity | PPV (5%) | NPV (5%) |
|---|---|---|---|---|---|
| Overall | 0.4274 | 0.9504 | 0.9302 | 0.4176 | 0.9972 |
| Single thoracic | 0.3601 | 1.000 | 0.9012 | 0.3475 | 1.000 |
| Single lumbar | 0.4095 | 0.9615 | 0.9128 | 0.3672 | 0.9978 |
| Multi-curve | 0.4274 | 0.9612 | 0.9302 | 0.4203 | 0.9978 |

PPV and NPV projected at 5% assumed screening prevalence.

### *D. Factor-Level Interpretability*

We tested whether the fixed KVT classifier relies on consistent kinematic evidence across AIS phenotypes on the external (DKC) cohort. Using attention-times-gradient attribution, we summarized, for each subgroup, mean temporal attention, domain-wise attribution mass over the gait cycle, and the top-ranked KKM factors.

Across general, single-thoracic, single-lumbar, and multi-curve strata (n = 282, 50, 26, and 206 AIS-positive participants, respectively), the two highest-attribution factors were the same: motion index 8 (horizontal position of the right ear) and skeleton index 142 (angle between upper extremities). Higher ranks were dominated by skeleton indices in all strata; motion indices appeared intermittently in the top 20. Top-factor magnitudes were similar across strata (for example, motion 8: 0.029 ± 0.002 thoracic vs 0.031 ± 0.003 lumbar vs 0.031 ± 0.001 multi) (**Fig. 6c-f**).

Domain dynamics (**Fig. 6b**) showed stable relative mass across the gait cycle, with temporal attention increasing toward later phases of the normalized sequence (**Fig. 6a**). These patterns support a stable, skeleton-oriented attributional template across curve locations. They are not causal claims: rankings describe correlates of a fixed supervised model, not interventions on individual factors.

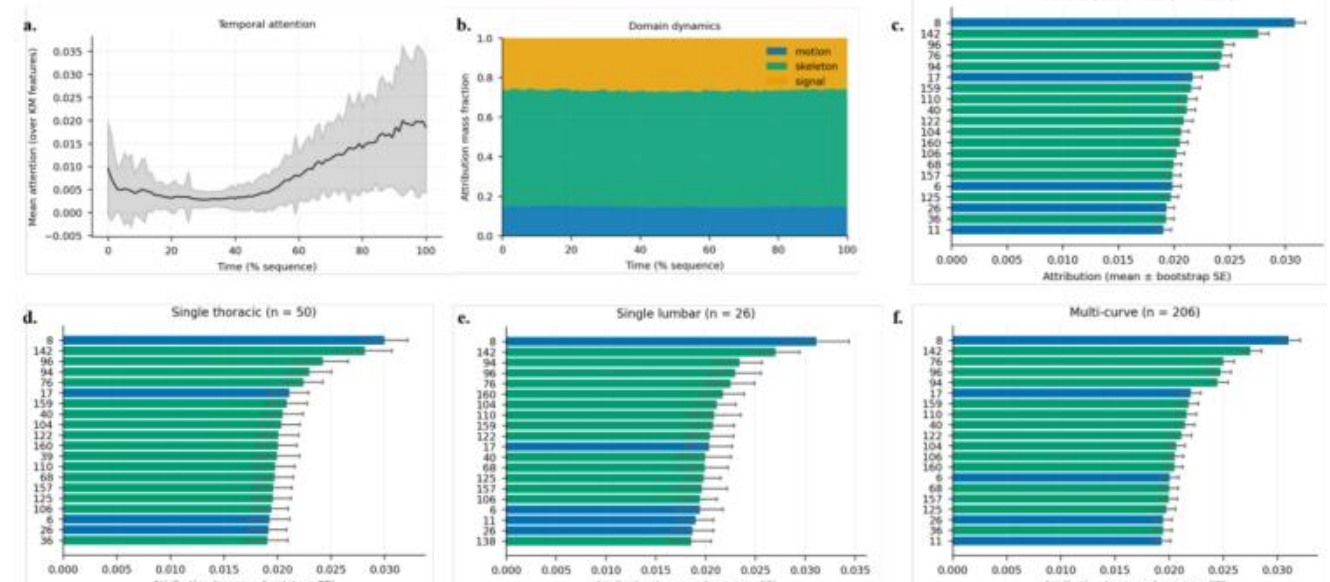

**Fig. 6.** Factor-level interpretability on the external test set. (a) Temporal attention. (b) Domain attribution vs. gait phase. (c)–(f) Top-20 factors by subgroup.

### *E. Input-level heterogeneity across datasets*

To characterize cross-site differences in model inputs, we compared KKM statistics between the internal development cohort (HKU-SZ, n = 820) and the external test cohort (n = 454 participant-level patches). The element-wise mean difference map (**Fig. 7a**) shows channel-specific shifts and stronger late-time structure. Per-patch Frobenius norms (**Fig. 7b**) are more concentrated externally, whereas the internal cohort has a broader distribution with heavier upper tails. Joint PCA of vectorized KKMs (**Fig. 7c**) places external samples in a compact neighbourhood, while internal samples span a wider region with isolated outliers; PC1 and PC2 explain 86.8% and 4.9% of variance, respectively.

Greater internal dispersion aligns with lower internal validation AUC versus external performance for comparable ViViT-based models (e.g., 0.787 vs. 0.961 under cohort shift), consistent with referral and acquisition heterogeneity on the hospital development arm rather than mismatched prediction targets across sites.

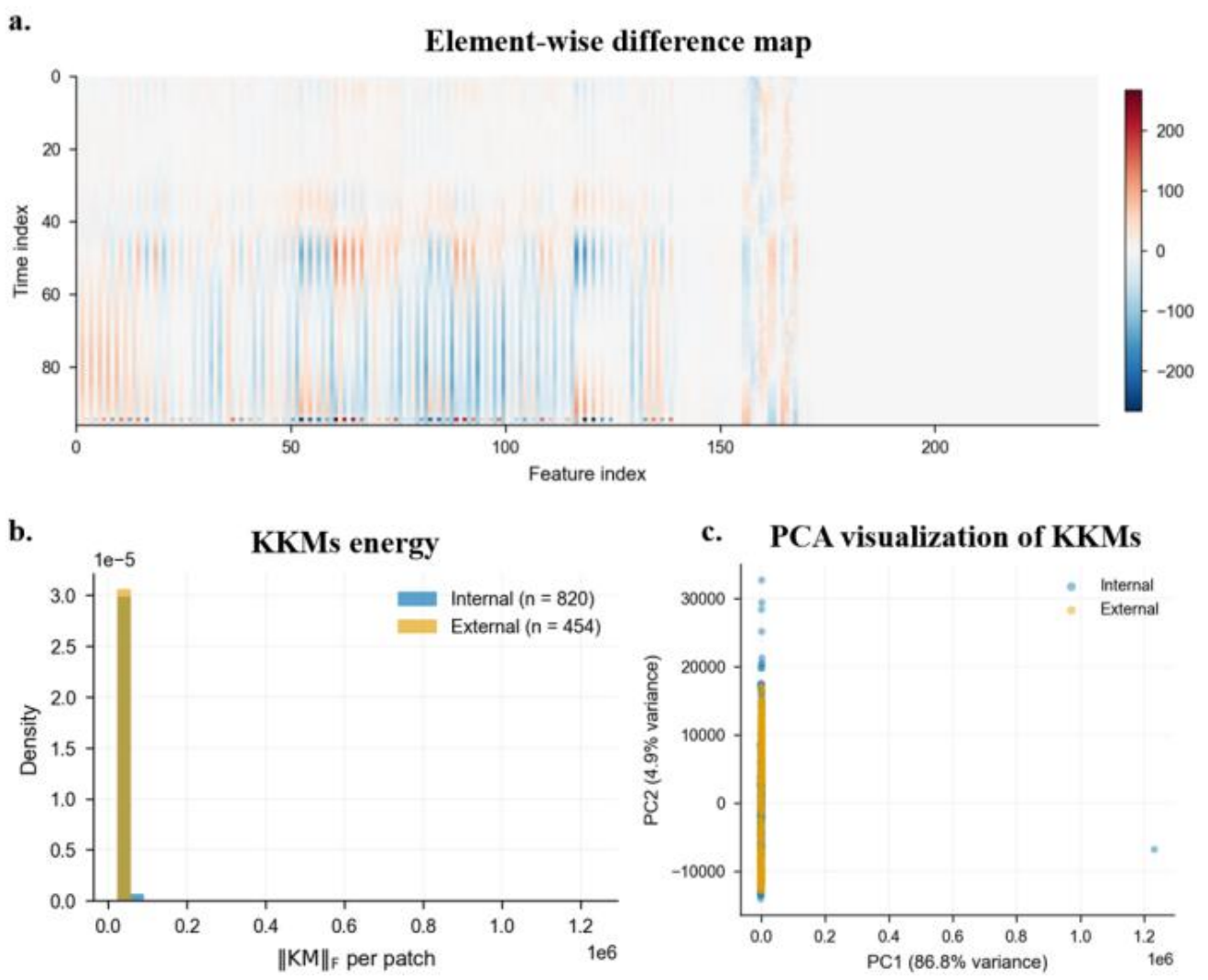

**Fig. 7.** Cross-cohort KKM geometry under acquisition shift. (a) Mean difference map. (b) Per-patch Frobenius norm (n = 820 vs. 454). (c) Joint PCA (PC1 86.8%; PC2 4.9%).

## VI. DISCUSSION

This work treats AIS screening as a pattern-analysis problem under weakly aligned multimodal inputs, such as monocular video, a derived structured time series, and template-based kinematic text produced on mismatched temporal grids and noise profiles. The core finding is that a fixed-index kinematic knowledge map provides a shared, auditable coordinate system in which modalities can be registered before fusion, yielding stronger and more interpretable models than pixel-only or late-concatenation baselines. On held-out external data, ScoliDetect (i) improves discrimination as structured kinematic mediation and latent-bottleneck fusion are added, (ii) remains stable across severity and curve-type strata under subject-level bootstrap inference, and (iii) yields consistent factor-level attributions across phenotypes. These outcomes follow from elevating representation design, fusion architecture, and indexed explanation to primary learning objectives rather than post-hoc visualization.

Deployment and intended use. ScoliDetect is intended as an assistive referral-triage aid under standardized monocular capture for school nurses, primary-care staff, and trained screening personnel—not as a substitute for clinical examination or radiography. Brief training and checklist-based capture were feasible in nine operators without retraining, supporting task shifting of video acquisition away from specialist technicians when supervision and protocol compliance are maintained.

The KKM encodes a testable inductive bias: periodic gait in screening can be approximated by a named factor basis that should be temporally aligned with video before cross-modal attention. Prespecified supervised ablations on the external cohort provide the primary evidence that KKM semantics complement pixels, that index-biased bidirectional cross-attention with latent-bottleneck aggregation outperforms late concatenation, and that auxiliary video-KKM InfoNCE, bottleneck fusion, cross-attention, and peak-anchored registration each contribute measurably. Trimodal contrastive pretraining was applied only after architecture selection, to initialize encoders rather than to justify fusion design. Backbone comparisons further suggest that the recipe transfers across Transformer video encoders under the fixed KVT protocol.

External ranking was stable across severity and curve-type subgroups, with pairwise ΔAUC contrasts near zero after multiplicity adjustment. Input-level geometry clarifies the internal and external performance gap: development KKMs are more dispersed, with heavier tails, than the compact external manifold observed under standardized school capture, consistent with referral heterogeneity on the hospital development arm rather than mismatched prediction targets. When acquisition compresses kinematic variability through protocol compliance, cross-modal integration becomes more reliable—a deployability lesson that argues for monitoring input dispersion before cross-site rollout. High external ROC-AUC should not be equated with standalone diagnostic utility: at a 5% assumed prevalence and sensitivity greater than 0.95, projected PPV remains modest. ScoliDetect is therefore intended as an assistive referral-triage aid under standardized monocular capture and brief checklist-based training was feasible in nine operators without model retraining.

Interpretability is structurally indexed but attributionally bounded. Because each KKM dimension maps to a predefined factor, readouts can be audited in biomechanical terms. However, gradient-based rankings describe correlates of a fixed classifier, not causal effects of interventions on individual factors. The same top skeleton-oriented factors across phenotypes suggest stable reasoning under fixed indexing rather than phenotype-specific saliency maps, but prospective reader or interventional studies are needed to test whether such readouts improve trust, audit, or failure detection in deployment.

Several limitations include (i) internal data dominated by a single hospital with many self-referred cases, limiting generalization of low internal performance to other unconstrained settings; (ii) rule-generated kinematic text rather than clinician narrative, so part of the multimodal gain may reflect template regularization; and (iii) gradient-based factor ranking at inference, which supports biomechanical audit but not standalone referral decisions. Feasibility testing with briefly trained operators supports standardized capture but not outcome effectiveness.

## VII. CONCLUSION

We presented SCOLIDETECT, a multimodal framework that aligns monocular gait video to a fixed-index kinematic knowledge map and template-based kinematic text through explicit temporal registration, index-biased video-KKM fusion, and late text concatenation at the screening head. On a multicenter cohort with patient-level partitioning and external school-screening validation, structured kinematic mediation and Perceiver-style latent-bottleneck fusion improved discrimination over unimodal and late-fusion baselines, remained stable across severity and curve-type subgroups, and supported factor-level explanations tied to model structure rather than post-hoc pixel saliency alone. Standardized capture was feasible with briefly trained non-clinical operators, positioning the system as referral triage under protocol rather than radiographic diagnosis.

Although developed for AIS screening, the framework applies broadly wherever video is paired with derived time series and template-based language, and where explanations must remain aligned with a predefined representational schema. Extending the same fixed-index contract to other periodic motion domains and testing it under prospective, multi-site deployment is a natural next step for pattern-analysis methods that must remain auditable as they cross acquisition boundaries.

## ACKNOWLEDGMENT

We acknowledge funding support by Sanming Project of Medicine in Shenzhen, China (No. SZSM 202211004); Shenzhen Science and Technology Program (No. KJZD20240903102759061); Shenzhen-Hong Kong Cooperation Zone for Technology and Innovation (No. HZQSWS-KCCYB-2024055). The funder played no role in study design, data collection, analysis and interpretation of data, or the writing of this manuscript.